\documentclass[10pt, a4paper]{article}
\usepackage{lrec}
\usepackage{multibib}
\newcites{languageresource}{Language Resources}
\usepackage{graphicx}
\usepackage{tabularx}
\usepackage{soul}

\usepackage{epstopdf}

\usepackage{hyperref}
\usepackage{xstring}

\usepackage{CJKutf8}
\usepackage{times}
\usepackage{latexsym}
\usepackage{pgf}
\usepackage{multirow}
\usepackage{booktabs}
\usepackage[title]{appendix}

\newcommand\enttype[1]{\texttt{#1}}

\title{BPEmb: Tokenization-free Pre-trained Subword Embeddings \\ in 275 Languages}

\name{Benjamin Heinzerling and Michael Strube}

\address{AIPHES, Heidelberg Institute for Theoretical Studies\\
	 Schloss-Wolfsbrunnenweg 35, 69118 Heidelberg, Germany \\
	 \{benjamin.heinzerling $\vert$ michael.strube\}@h-its.org}

\abstract{
	We present \emph{BPEmb}, a collection of pre-trained subword unit embeddings in 275 languages, based on Byte-Pair Encoding (BPE). In an evaluation using fine-grained entity typing as testbed, BPEmb performs competitively, and for some languages better than alternative subword approaches, while requiring vastly fewer resources and no tokenization. BPEmb is available at \url{https://github.com/bheinzerling/bpemb}. \\ \newline  }

\begin{document}

\begin{CJK}{UTF8}{min}

\maketitleabstract

\section{Introduction}

Learning good representations of rare words or words not seen during training at all is a difficult challenge in natural language processing.
As a makeshift solution, systems have typically replaced such words with a generic \emph{UNK} token. 
Recently, based on the assumption that a word's meaning can be reconstructed from its parts, several subword-based methods have been proposed to deal with the unknown word problem: character-based recurrent neural networks (RNN) \cite{luong16}, character-based convolutional neural networks (CNN) \cite{chiu16}, word embeddings enriched with subword information (FastText) \cite{bojanowski17}, and byte-pair encoding (BPE) \cite{sennrich16}, among others. While pre-trained FastText embeddings are publicly available, embeddings for BPE units are commonly trained on a per-task basis (e.g. a specific language pair for machine-translation) and not published for general use.

In this work we present \emph{BPEmb}, a collection of pre-trained subword embeddings in 275 languages, and make the following contributions:
\begin{itemize}
\item We publish BPEmb, a collection of pre-trained byte-pair embeddings in 275 languages;
\item We show the utility of BPEmb in a fine-grained entity typing task; and
\item We show that BPEmb performs as well as, and for some languages better than, alternative approaches while being more compact and requiring no tokenization.
\end{itemize}

\section{BPEmb: Byte-pair Embeddings}

Byte Pair Encoding is a variable-length encoding that views text as a sequence of symbols and iteratively merges the most frequent symbol pair into a new symbol. E.g., encoding an English text might consist of first merging the most frequent symbol pair \emph{t h} into a new symbol \emph{th}, then merging the pair \emph{th e} into \emph{the} in the next iteration, and so on. The number of merge operations $o$ determines if the resulting encoding mostly creates short character sequences (e.g. $o = 1000$) or if it includes symbols for many frequently occurring words, e.g. $o = 30,000$ (cf. Table~\ref{tbl:exops}).
Since the BPE algorithm works with any sequence of symbols, it requires no preprocessing and can be applied to untokenized text.

We apply BPE\footnote{We use the SentencePiece BPE implementation: \url{https://github.com/google/sentencepiece}.} to all Wikipedias\footnote{%
We extract text from Wikipedia articles with WikiExtract (\url{http://attardi.github.io/wikiextractor}), lowercase all characters where applicable and map all digits to zero.}
of sufficient size with various $o$ and pre-train embeddings for the resulting BPE symbol using GloVe \cite{pennington14}, resulting in byte-pair embeddings for 275 languages.
To allow studying the effect the number of BPE merge operations and of the embedding dimensionality, we provide embeddings for 1000, 3000, 5000, 10000, 25000, 50000, 100000 and 200000 merge operations, with dimensions 25, 50, 100, 200, and 300. 

\begin{table*}
	\centering
	\begin{tabular}{r|l}
		\toprule
		Merge ops\enspace&\enspace Byte-pair encoded text \\
		\midrule
		5000\enspace &\enspace 豊\enspace\enspace 田\enspace 駅\enspace (\enspace と\enspace よ\enspace だ\enspace え\enspace き\enspace )\enspace は\enspace 、\enspace 東京都\enspace 日\enspace 野\enspace 市\enspace 豊\enspace 田\enspace 四\enspace 丁目\enspace にある\enspace \\
		10000\enspace &\enspace 豊\enspace 田\enspace 駅\enspace (\enspace と\enspace よ\enspace だ\enspace えき\enspace )\enspace は\enspace 、\enspace 東京都\enspace 日\enspace 野市\enspace 豊\enspace 田\enspace 四\enspace 丁目にある\enspace \\
		25000\enspace &\enspace 豊\enspace 田駅\enspace (\enspace とよ\enspace だ\enspace えき\enspace )\enspace は\enspace 、\enspace 東京都\enspace 日\enspace 野市\enspace 豊田\enspace 四\enspace 丁目にある\enspace \\
		50000\enspace &\enspace 豊\enspace 田駅\enspace (\enspace とよ\enspace だ\enspace えき\enspace )\enspace は\enspace 、\enspace 東京都\enspace 日\enspace 野市\enspace 豊田\enspace 四丁目にある\enspace \\
		Tokenized\enspace &\enspace 豊田\enspace 駅\enspace （\enspace と\enspace よ\enspace だ\enspace え\enspace き\enspace ）\enspace は\enspace 、\enspace 東京\enspace 都\enspace 日野\enspace 市\enspace 豊田\enspace 四\enspace 丁目\enspace に\enspace ある\enspace \\
		\midrule
		10000\enspace &\enspace 豐\enspace 田\enspace 站\enspace 是\enspace 東\enspace 日本\enspace 旅\enspace 客\enspace 鐵\enspace 道\enspace (\enspace JR\enspace 東\enspace 日本\enspace )\enspace 中央\enspace 本\enspace 線\enspace 的\enspace 鐵路\enspace 車站\enspace \\
		25000\enspace &\enspace 豐田\enspace 站是\enspace 東日本旅客鐵道\enspace (\enspace JR\enspace 東日本\enspace )\enspace 中央\enspace 本\enspace 線的鐵路車站\enspace \\
		50000\enspace &\enspace 豐田\enspace 站是\enspace 東日本旅客鐵道\enspace (\enspace JR\enspace 東日本\enspace )\enspace 中央\enspace 本線的鐵路車站\enspace \\
		Tokenized\enspace &\enspace 豐田站\enspace 是\enspace 東日本\enspace 旅客\enspace 鐵道\enspace （\enspace JR\enspace 東日本\enspace ）\enspace 中央本線\enspace 的\enspace 鐵路車站\enspace \\
		\midrule
		1000\enspace &\enspace to\enspace y\enspace od\enspace a\enspace \_station\enspace is\enspace \_a\enspace \_r\enspace ail\enspace way\enspace \_station\enspace \_on\enspace \_the\enspace \_ch\enspace \={u}\enspace \={o}\enspace \_main\enspace \_l\enspace ine\\
		3000\enspace &\enspace to\enspace y\enspace od\enspace a\enspace \_station\enspace \_is\enspace \_a\enspace \_railway\enspace \_station\enspace \_on\enspace \_the\enspace \_ch\enspace \={u}\enspace \={o}\enspace \_main\enspace \_line\\
		10000\enspace &\enspace toy\enspace oda\enspace \_station\enspace \_is\enspace \_a\enspace \_railway\enspace \_station\enspace \_on\enspace \_the\enspace \_ch\enspace \={u}\enspace \={o}\enspace \_main\enspace \_line\\
		50000\enspace &\enspace toy\enspace oda\enspace \_station\enspace \_is\enspace \_a\enspace \_railway\enspace \_station\enspace \_on\enspace \_the\enspace \_ch\={u}\enspace \={o}\enspace \_main\enspace \_line\enspace \\
		100000\enspace &\enspace toy\enspace oda\enspace \_station\enspace \_is\enspace \_a\enspace \_railway\enspace \_station\enspace \_on\enspace \_the\enspace \_ch\={u}\={o}\enspace \_main\enspace \_line\enspace \\
		Tokenized\enspace &\enspace toyoda\enspace station\enspace is\enspace a\enspace railway\enspace station\enspace on\enspace the\enspace ch\={u}\={o}\enspace main\enspace line\enspace \\
		\bottomrule
	\end{tabular}
	\caption{Effect of the number of BPE merge operations on the beginning of the Japanese (top), Chinese (middle), and English (bottom) Wikipedia article \textsc{Toyoda\_Station}.
	Since BPE is based on frequency, the resulting segmentation is often, but not always meaningful.
	E.g. in the Japanese text, 豊 (toyo) 
		and 田 (ta) are correctly merged into 豊田 (Toyoda, a Japanese city) in the second occurrence,
	but the first 田 is first merged with 駅 (eki, \emph{train station}) into the meaningless 田駅 (ta-eki).
	}
	\label{tbl:exops}
\end{table*}

\section{Evaluation: Comparison to FastText and Character Embeddings}

To evaluate the quality of BPEemb we compare to FastText, a state-of-the-art approach that combines embeddings of tokens and subword units, as well as to character embeddings.

\noindent\textbf{FastText} enriches word embeddings with subword information by additionally learning embeddings for character n-grams. A word is then represented as the sum of its associated character n-gram embeddings including. In practice, representations of unknown word are obtained by adding the embeddings of their constituting character 3- to 6-grams. We use the pre-trained embeddings provided by the authors.\footnote{\url{https://github.com/facebookresearch/fastText}}

\noindent\textbf{Character embeddings.} In this setting, mentions are represented as sequence of the character unigrams\footnote{We also studied character bigrams and trigrams. Results were similar to unigrams and are omitted for space.} they consist of. During training, character embeddings are learned for the $k$ most frequent characters.

\textbf{Fine-grained entity typing.} Following \newcite{schutze17}, we use fine-grained entity typing as test bed for comparing subword approaches. This is an interesting task for subword evaluation, since many rare, long-tail entities do not have good representations in common token-based pre-trained embeddings such as word2vec or GloVe. 
Subword-based models are a promising approach to this task, since morphology often reveals the semantic category of unknown words: The suffix \emph{-shire} in \emph{Melfordshire} indicates a location or city, and the suffix \emph{-osis} in \emph{Myxomatosis} a sickness. Subword methods aim to allow this kind of inference by learning representations of subword units (henceforth: SUs) such as character ngrams, morphemes, or byte pairs.

\noindent\textbf{Method.} Given an entity mention $m$ such as \emph{Melfordshire}, our task is to assign one or more of the 89 fine-grained entity types proposed by \newcite{gillick14}, in this case \enttype{/location} and \enttype{/location/city}. To do so, we first obtain a subword representation
$$s = SU(m) \in R^{l \times d}$$
by applying one of the above SU transformations resulting in a SU sequence of length $l$ and then looking up the corresponding SU embeddings with dimensionality $d$. Next, $s$ is encoded into a one-dimensional vector representation
$$v = A(s) \in R^d$$ by an encoder $A$. In this work the encoder architecture is either averaging across the SU sequence, an LSTM, or a CNN. Finally, the prediction $y$ is:
$$y = \frac{1}{1 + exp(-v)}$$ \cite{shimaoka17}.

\vspace{1.1ex}

\noindent\textbf{Data.} We obtain entity mentions from Wikidata \cite{vrandevcic14} and their entity types by mapping to Freebase \cite{bollacker08}, resulting in 3.4 million English\footnote{
Numbers for other languages omitted for space.} instances like (\emph{Melfordshire}: \enttype{/location},\enttype{/location/city}). Train and test set are random subsamples of size 80,000 and 20,000 or a proportionally smaller split for smaller Wikipedias.
In addition to English, we report results for a) the five languages having the largest Wikipedias as measured by textual content; b) Chinese and Japanese, i.e. two high-resource languages without tokenization markers; and c) eight medium- to low-resource Asian languages.

\noindent\textbf{Experimental Setup.} We evaluate entity typing performance with the average of strict, loose micro, and loose macro precision \cite{ling12}. For each combination of SU and encoding architecture, we perform a Tree-structured Parzen Estimator hyper-parameter search \cite{bergstra11} with at least 1000 hyper-parameter search trials (English, at least 50 trials for other languages) and report score distributions \cite{reimers17}. See Table~\ref{tbl:hparams} for hyper-parameter ranges.

\section{Results and Discussion}

\begin{figure*}
	\scalebox{.95}[.95]{
	\input{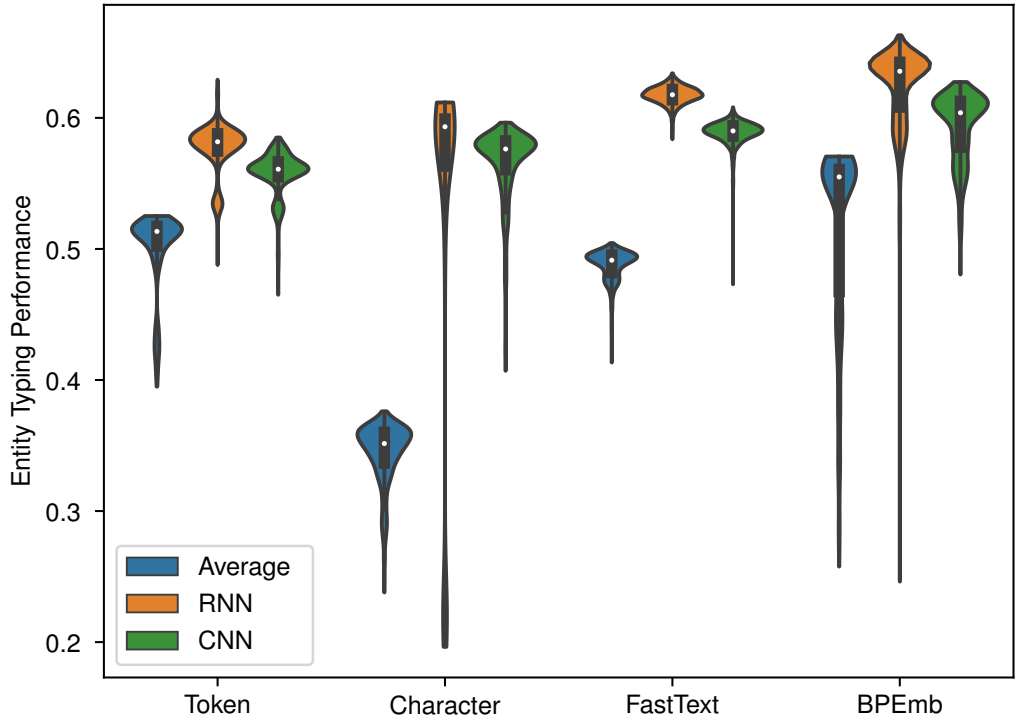}
	}
	\vspace{-3.5ex}
	\caption{English entity typing performance of subword embeddings across different architectures. This violin plot shows smoothed distributions of the scores obtained during hyper-parameter search. White points represent medians, boxes quartiles. Distributions are cut to reflect highest and lowest scores.}
	\label{fig:results}
\end{figure*}

\subsection{Subwords vs. Characters vs. Tokens}

Figure~\ref{fig:results} shows our main result for English: score distributions of 1000+ trials for each SU and architecture. Token-based results using two sets of pre-trained embeddings \cite{mikolov13b,pennington14} are included for comparison.

\textbf{Subword units.} BPEmb outperforms all other subword units across all architectures (BPE-RNN mean score $0.624 \pm 0.029$, max. $0.65$). FastText performs slightly worse (FastText-RNN mean $0.617 \pm 0.007$, max. $0.63$)\footnote{Difference to BPEmb significant, $p<0.001$, Approximate Randomization Test.}, even though the FastText vocabulary is much larger than the set of BPE symbols.

\begin{figure*}
	\centering
	\scalebox{1}[1.061]{
	\includegraphics[width=0.5\linewidth]{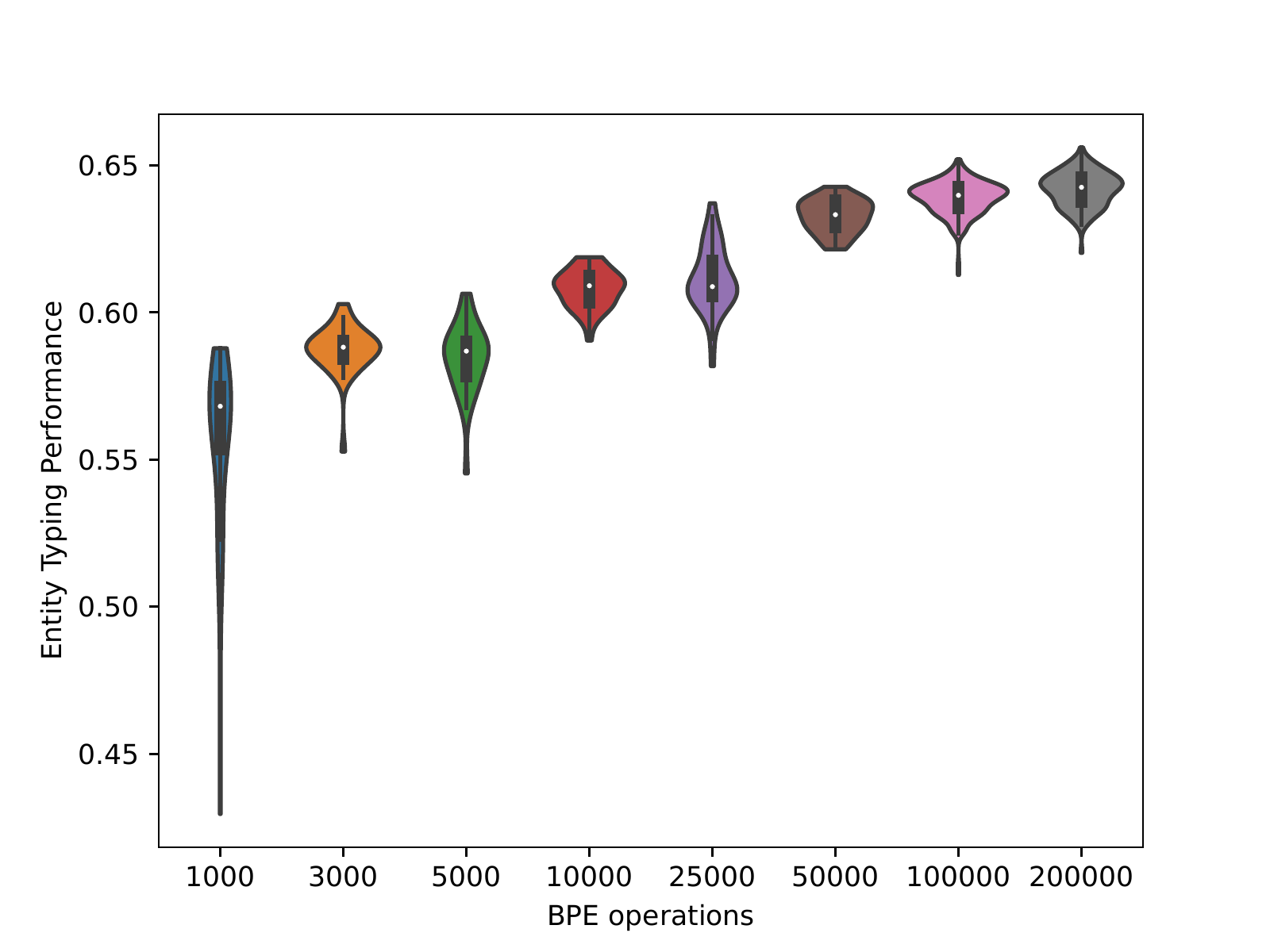}
	\hspace{-0.85em}
	\includegraphics[width=0.5\linewidth]{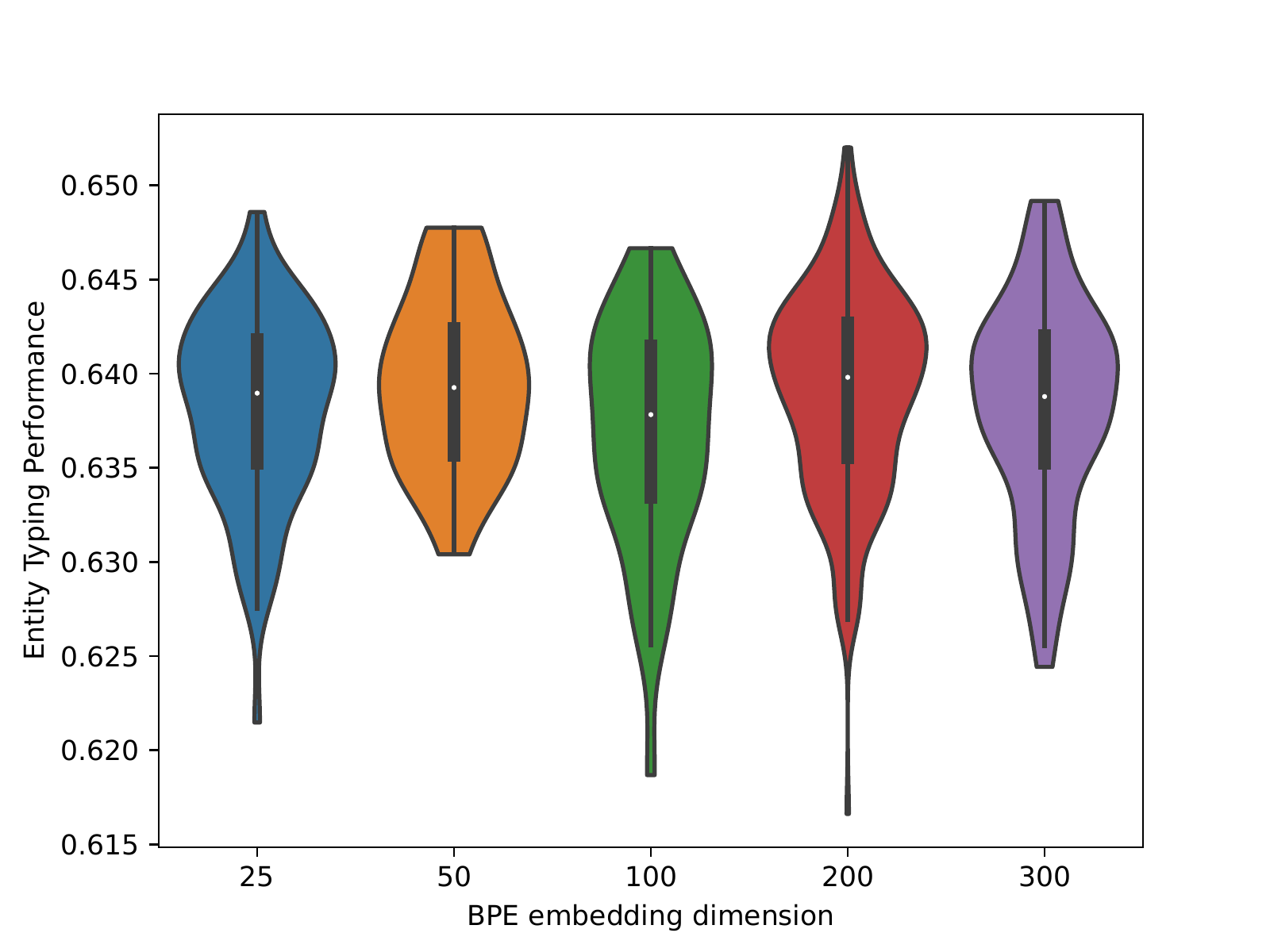}
	}
	\caption{Impact of the number of BPE merge operations (left) and embedding dimension (right) on English entity typing.}
	\label{fig:opsdim}
\end{figure*}

BPEmb performs well with low embedding dimensionality Figure~\ref{fig:opsdim}, right) and can match FastText with a fraction of its memory footprint (6 GB for FastText's 3 million embeddings with dimension $300$ vs 11 MB for 100k BPE embeddings (Figure~\ref{fig:opsdim}, left) with dimension $25$.). As both FastText and BPEmb were trained on the same corpus (namely, Wikipedia), these results suggest that, for English, the compact BPE representation strikes a better balance between learning embeddings for more frequent words and relying on compositionality of subwords for less frequent ones.

FastText performance shows the lowest variance, i.e., it robustly yields good results across many different hyper-parameter settings. In contrast, BPEmb and character-based models show higher variance, i.e., they require more careful hyper-parameter tuning to achieve good results.

\textbf{Architectures.} Averaging a mention's associated embeddings is the worst architecture choice. This is expected for character-based models, but somewhat surprising for token-based models, given the fact that averaging is a common method for representing mentions in tasks such as entity typing \cite{shimaoka17} or coreference resolution \cite{clarkkevin16a}. RNNs perform slightly better than CNNs, at the cost of much longer training time.

\begin{table}
	\vspace{1.4ex}
	\centering
	\begin{tabular}{lrrr}
		\toprule
		Language & FastText & BPEmb & $\Delta$ \\
		\midrule
		English & 62.9 & \textbf{65.4} & 2.5 \\
		German & 65.5 & \textbf{66.2} & 0.7 \\
		Russian & \textbf{71.2} & 70.7 & -0.5 \\
		French & \textbf{64.5} & 63.9 & -0.6 \\
		Spanish & \textbf{66.6} & 66.5 & -0.1 \\
		\midrule
		Chinese & 71.0 & \textbf{72.0} & 1.0 \\
		Japanese & \textbf{62.3} & 61.4 & -0.9 \\
		\midrule
		Tibetan & 37.9 & \textbf{41.4} & 3.5 \\
		Burmese & \textbf{65.0} & 64.6 & -0.4 \\
		Vietnamese & 81.0 & 81.0 & 0.0 \\
		Khmer & \textbf{61.5} & 52.6 & -8.9 \\
		Thai & 63.5 & \textbf{63.8} & 0.3 \\
		Lao & 44.9 & \textbf{47.0} & 2.1 \\
		Malay & 75.9 & \textbf{76.3} & 0.4 \\
		Tagalog & \textbf{63.4} & 62.6 & -1.2 \\
		\bottomrule
	\end{tabular}
	\vspace{1.4ex}
	\caption{Entity typing scores for five high-resource languages (top), two high-resource languages without explicit tokenization, and eight medium- to low-resource Asian languages (bottom).}
	\label{tbl:multires}
\end{table}

\subsection{Multilingual Analysis}

Table~\ref{tbl:multires} compares FastText and BPEmb across various languages. For high-resource languages (top) both approaches perform equally, with the exception of BPEmb giving a significant improvement for English. For high resources languages without explicit tokenization (middle), byte-pair encoding appears to yield a subword segmentation which gives performance comparable to the results obtained when using FastText with pre-tokenized text\footnote{Tokenization for Chinese was performed with Stanford CoreNLP \cite{manning14} and for Japanese with Kuromoji (\url{https://github.com/atilika/kuromoji}).}.

Results are more varied for mid- to low-resource Asian languages (bottom), with small BPEmb gains for Tibetan and Lao. The large performance degradation for Khmer appears to be due to inconsistencies in the handling of unicode control characters between different software libraries used in our experiments and have a disproportionate effect due to the small size of the Khmer Wikipedia.

\section{Limitations}

Due to limited computational resources, our evaluation was performed only for a few of the 275 languages provided by BPEemb.
While our experimental setup allows a fair comparison between FastText and BPEmb through extensive hyper-parameter search, it is somewhat artificial, since it disregards context. For example, \emph{Myxomatosis} in the phrase \emph{Radiohead played Myxomatosis} has the entity type \texttt{/other/music}, which can be inferred from the contextual music group and the predicate \emph{plays}, but this ignored in our specific setting. How our results transfer to other tasks requires further study.

\section{Replicability}

All data used in this work is freely and publicly available. Code to replicate our experiments will be released upon publication at \url{https://github.com/bheinzerling/bpemb}.

\section{Conclusions}

We presented BPEmb, a collection of subword embeddings trained on Wikipedias in 275 languages. Our evaluation showed that BPEmb performs as well as, and for some languages, better than other subword-based approaches. BPEmb requires no tokenization and is orders of magnitudes smaller than alternative embeddings, enabling potential use under resource constraints, e.g. on mobile devices. 

\section*{Acknowledgments}

This work has been supported by the German Research Foundation as part of the Research Training Group ``Adaptive Preparation of Information from Heterogeneous Sources'' (AIPHES) under grant No. GRK 1994/1, and partially funded by the Klaus Tschira Foundation, Heidelberg, Germany. 

\begin{table}[h]
\centering
\resizebox{\columnwidth}{!}{%
	\begin{tabular}{lll}
	\toprule
	\textbf{Unit} & \textbf{Hyper-parameter} & \textbf{Space} \\
	\midrule
	Token & embedding type & GloVe, word2vec \\
	\midrule
	\multirow{2}{*}{Character} & vocabulary size & 50, 100, 200, 500, 1000 \\
	 & embedding dimension & 10, 25, 50, 75, 100 \\
	\midrule
	FastText & - & - \\
	\midrule
	\multirow{3}{*}{BPE} & merge operations & 1k, 3k, 5k, 10k, 25k \\
	 & & 50k, 10k, 200k \\
	 & embedding dimension & 25, 50, 100, 200, 300 \\

	\\[1ex]

	\toprule
	\textbf{Architecture} & \textbf{Hyper-parameter} & \textbf{Space} \\
	\midrule
	\multirow{4}{*}{RNN} & hidden units & 100, 300, 500, 700, \\
	 & & 1000, 1500, 2000 \\
	 & layers & 1, 2, 3 \\
	 & RNN dropout & 0.0, 0.1, 0.2, 0.3, 0.4, 0.5 \\
	 & output dropout & 0.0, 0.1, 0.2, 0.3, 0.4, 0.5 \\
	\midrule
	\multirow{7}{*}{CNN} & filter sizes & (2), (2, 3), (2, 3, 4), \\
		& & (2, 3, 4, 5), (2, 3, 4, 5, 6), \\
		& & (3), (3, 4), (3, 4, 5), (3, 4, 5, 6), \\
		& & (4), (4, 5), (4, 5, 6), (5), (5, 6), (6) \\
	 & number of filters & 25, 50, 100, 200, \\
	 & & 300, 400, 500, 600, 700 \\
	 & output dropout & 0.0, 0.1, 0.2, 0.3, 0.4, 0.5 \\
	\midrule
	Average & output dropout & 0.0, 0.1, 0.2, 0.3, 0.4, 0.5 \\
	\bottomrule
	\end{tabular}
}
\caption{Subword unit (top) and architecture (bottom) hyper-parameter space searched.}
\label{tbl:hparams}
\end{table}

\section{Bibliographical References}
\label{main:ref}

\bibliographystyle{lrec}
\bibliography{lit,add}

\begin{thebibliography}{}

\bibitem[\protect\citename{Bergstra \bgroup et al.\egroup }2011]{bergstra11}
Bergstra, J.~S., Bardenet, R., Bengio, Y., and K{\'e}gl, B.
\newblock (2011).
\newblock Algorithms for hyper-parameter optimization.
\newblock In {\em Advances in Neural Information Processing Systems}, pages
  2546--2554.

\bibitem[\protect\citename{Bojanowski \bgroup et al.\egroup
  }2017]{bojanowski17}
Bojanowski, P., Grave, E., Joulin, A., and Mikolov, T.
\newblock (2017).
\newblock Enriching word vectors with subword information.
\newblock {\em Transactions of the Association for Computational Linguistics},
  5:135--146.

\bibitem[\protect\citename{Bollacker \bgroup et al.\egroup }2008]{bollacker08}
Bollacker, K., Evans, C., Paritosh, P., Sturge, T., and Taylor, J.
\newblock (2008).
\newblock Freebase: A collaboratively created graph database for structuring
  human knowledge.
\newblock In {\em Proceedings of the 2008 ACM SIGMOD International Conference
  on Management of Data, {\em Vancouver, B.C., Canada, 10--12 June 2008}},
  pages 1247--1250.

\bibitem[\protect\citename{Chiu and Nichols}2016]{chiu16}
Chiu, J. and Nichols, E.
\newblock (2016).
\newblock Named entity recognition with bidirectional {LSTM}-{CNNs}.
\newblock {\em Transactions of the Association for Computational Linguistics},
  4:357--370.

\bibitem[\protect\citename{Clark and Manning}2016]{clarkkevin16a}
Clark, K. and Manning, C.~D.
\newblock (2016).
\newblock Improving coreference resolution by learning entity-level distributed
  representations.
\newblock In {\em Proceedings of the 54th Annual Meeting of the Association for
  Computational Linguistics (Volume 1: Long Papers), {\em Berlin, Germany,
  7--12 August 2016}}.

\bibitem[\protect\citename{{Gillick} \bgroup et al.\egroup }2014]{gillick14}
{Gillick}, D., {Lazic}, N., {Ganchev}, K., {Kirchner}, J., and {Huynh}, D.
\newblock (2014).
\newblock {Context-Dependent Fine-Grained Entity Type Tagging}.
\newblock {\em ArXiv e-prints}, December.

\bibitem[\protect\citename{Ling and Weld}2012]{ling12}
Ling, X. and Weld, D.~S.
\newblock (2012).
\newblock Fine-grained entity recognition.
\newblock In {\em AAAI}.

\bibitem[\protect\citename{Luong and Manning}2016]{luong16}
Luong, M.-T. and Manning, C.~D.
\newblock (2016).
\newblock Achieving open vocabulary neural machine translation with hybrid
  word-character models.
\newblock In {\em Proceedings of the 54th Annual Meeting of the Association for
  Computational Linguistics (Volume 1: Long Papers)}, pages 1054--1063, Berlin,
  Germany, August. Association for Computational Linguistics.

\bibitem[\protect\citename{Manning \bgroup et al.\egroup }2014]{manning14}
Manning, C.~D., Surdeanu, M., Bauer, J., Finkel, J., Bethard, S.~J., and
  McClosky, D.
\newblock (2014).
\newblock The {Stanford} {CoreNLP} natural language processing toolkit.
\newblock In {\em Proceedings of 52nd Annual Meeting of the Association for
  Computational Linguistics: System Demonstrations}, pages 55--60. Association
  for Computational Linguistics.

\bibitem[\protect\citename{Mikolov \bgroup et al.\egroup }2013]{mikolov13b}
Mikolov, T., Chen, K., Corrado, G., and Dean, J.
\newblock (2013).
\newblock Efficient estimation of word representations in vector space.
\newblock In {\em Proceedings of the ICLR 2013 Workshop Track}.

\bibitem[\protect\citename{Pennington \bgroup et al.\egroup
  }2014]{pennington14}
Pennington, J., Socher, R., and Manning, C.
\newblock (2014).
\newblock Glove: Global vectors for word representation.
\newblock In {\em Proceedings of the 2014 Conference on Empirical Methods in
  Natural Language Processing (EMNLP)}, pages 1532--1543, Doha, Qatar, October.
  Association for Computational Linguistics.

\bibitem[\protect\citename{Reimers and Gurevych}2017]{reimers17}
Reimers, N. and Gurevych, I.
\newblock (2017).
\newblock Reporting score distributions makes a difference: Performance study
  of lstm-networks for sequence tagging.
\newblock {\em CoRR}, abs/1707.09861.

\bibitem[\protect\citename{Sch\"{u}tze}2017]{schutze17}
Sch\"{u}tze, H.
\newblock (2017).
\newblock Nonsymbolic text representation.
\newblock In {\em Proceedings of the 15th Conference of the European Chapter of
  the Association for Computational Linguistics: Volume 1, Long Papers}, pages
  785--796, Valencia, Spain, April. Association for Computational Linguistics.

\bibitem[\protect\citename{Sennrich \bgroup et al.\egroup }2016]{sennrich16}
Sennrich, R., Haddow, B., and Birch, A.
\newblock (2016).
\newblock Neural machine translation of rare words with subword units.
\newblock In {\em Proceedings of the 54th Annual Meeting of the Association for
  Computational Linguistics (Volume 1: Long Papers)}, pages 1715--1725, Berlin,
  Germany, August. Association for Computational Linguistics.

\bibitem[\protect\citename{Shimaoka \bgroup et al.\egroup }2017]{shimaoka17}
Shimaoka, S., Stenetorp, P., Inui, K., and Riedel, S.
\newblock (2017).
\newblock Neural architectures for fine-grained entity type classification.
\newblock In {\em Proceedings of the 15th Conference of the European Chapter of
  the Association for Computational Linguistics: Volume 1, Long Papers}, pages
  1271--1280, Valencia, Spain, April. Association for Computational
  Linguistics.

\bibitem[\protect\citename{Vrande{\v{c}}i{\'c} and
  Kr{\"o}tzsch}2014]{vrandevcic14}
Vrande{\v{c}}i{\'c}, D. and Kr{\"o}tzsch, M.
\newblock (2014).
\newblock Wikidata: a free collaborative knowledgebase.
\newblock {\em Communications of the ACM}, 57(10):78--85.

\end{thebibliography}

\clearpage\end{CJK}

\end{document}